# Implications of Multi-Word Expressions on English to Bharti Braille Machine Translation

Nisheeth Joshi[1,2], Pragya Katyayan[1,2]
[1]Department of Computer Science, Banasthali Vidyapith, Rajasthan, India
[2]Centre for Artificial Intelligence, Banasthali Vidyapith, Rajasthan
nisheeth.joshi@rediffmail.com, pragya.katyayan@outlook.com

*Abstract*— In this paper, we have shown the improvement of English to Bharti Braille machine translation system. We have shown how we can improve a baseline NMT model by adding some linguistic knowledge to it. This was done for five language pairs where English sentences were translated into five Indian languages and then subsequently to corresponding Bharti Braille. This has been demonstrated by adding a sub-module for translating multi-word expressions. The approach shows promising results as across language pairs, we could see improvement in the quality of NMT outputs. The least improvement was observed in English-Nepali language pair with 22.08% and the most improvement was observed in the English-Hindi language pair with 23.30%

*Keywords—multiwords, neural machine translation, bharti braille*

## I. INTRODUCTION

A multiword expression (MWE) is a phrase or group of words that function as a single unit and convey a specific meaning that cannot be derived from the meanings of its individual words. Examples of MWEs include idioms, phrasal verbs, and compound nouns. MWEs are a common challenge in natural language processing, as they often require special handling to ensure accurate translation or understanding. In NLP tasks such as machine translation and information extraction, proper recognition and processing of MWEs can greatly improve the performance and accuracy of the systems.

Multi-word expressions (MWEs) can pose a challenge in machine translation. MWEs often have idiomatic meanings that cannot be derived from the meanings of individual words, making them difficult to translate correctly. Additionally, MWEs may have different forms and meanings across different languages, which can lead to incorrect translations if not handled properly. To address these challenges, researchers have developed several methods for handling MWEs in machine translation, such as using parallel corpora, specialized lexicons, and statistical models. These methods aim to capture the meaning and variation of MWEs across languages, and to integrate them into the machine translation process to improve the quality of translations. Proper handling of MWEs can greatly enhance the accuracy and fluency of machine translation outputs.

Bharti Braille is a form of Braille script used in India to write the languages of India in Braille script. It is based on the standard Braille script but includes additional characters to represent the unique sounds and characters found in Indian languages. It has been widely adopted in India and is used in schools and other educational institutions to teach visually impaired students. It has also been used to create Braille books and other materials in Indian languages, making it easier for visually impaired people to access literature and other written materials in their native language.

## II. LITERATURE SURVEY

Lewis et al. [1] have developed acceptable performance items for demonstarting competence I literary Braille. Hong et al. (2012) devised an alternative method for Braille notetakers for visually impaired people in the form of a dedicated notetaking device for Braille. Herzberg and Rosenblum [2] provided an analysis of 107 mathematical worksheets prepared for visually impaired students on the basis of accuracy. Al-Salman et al. [3] proposed new method for Braille image segmentation by utilizing between-class variance method with gamma distribution given by Otsu. Their method had two main segments: first, finding optimally estimated threshold using the variance technique with gamma distribution mixture and second, using the optimal thresholds for braille image segmentation.

Authman and Jebr [4] have described a new technique for identifying Braille cells in a single-sided Braille document in Arabic. Their optical Braille Recognition system for Arabic performs two tasks: first is to identify printed Braille cells and second is converting them as normal text. Padmavathi et al. [4] have proposed a method to convert a scanned document of Braille into completely readable text format. Braille documents were pre-processed to reduce noise and enhance the dots. Dots were then extracted after Braille cells were segmented and was changed in a sequence of numbers. It is then mapped to correct alphabets of the language (English, Hindi or Tamil) and is read aloud with the help of a speech synthesizer. They also gave a method of typing Braille via the numeric pad on keyboard.

Abualkishik and Omar [5] have introduced a Quran Braille Translator which could translate verses from Quran to Braille. They used extended finite state machine (EFSM) for finding reciting rules of Quran and Markov algorithm to translate reciting rules along with Quran text to appropriate Braille code. Oda et al. [6] adapted ML for NLP to improve word segmentation for web-based automatic translation program for braille. They created statistical models using a SVM-based general purpose chunker. Hossain et al. [7] have identified rules and conventions for Bangla Braille translation based on rules. They proposed a DFA based computational model for MT, which gave acceptable translations. The results were tested by members of visually impaired community.

Al-Salman et al. [8] have built a Braille copier machine which produced Braille documents in their respective languages. The machine worked as both copier as well as printing system using optical recognition and techniques from image processing. Yamaguchi et al. [9] have highlighted the problem of accuracy while translating technical notations to Braille. To solve this problem, they have developed an assistive solution for people from STEM background who are not capable of printing. Damit et al. [10] have mediated a new way of interlinking keyboard inputs from translates to commonly used Braille characters. This enabled visually blessed people to interact with visually impaired people.

Rupanagudi et al. [11] introduced a new technique of translating Kannada braille to Kannada language. They devised a new algorithm to segment Braille dots and identified characters. Choudhary et al. [12] have suggested a new approach for supporting communication amongst deaf-blind people. The technique included the use of smart gloves capable of translating Braille alphabets and can communicate the message via SMS. Due to this the user can convey simple messages using touch sensors. Guerra et al. [13] have developed a proto-type using Java which can convert Braille text to digital text. Jariwala and Patel [14] have developed tool for translation of Gujarati, English and Hindi text to Braille and save it as a data file which can be directly printed via embosser.

Saxena et al. [15] have provided a real-time solution (hardware and software) for helping blind people. They developed a Braille hand glove which helped in communication for sending and receiving messages in real time. Nam and Min [16] have developed a music braille converted capable of converting the musical notations such as octaves, key signature, tie repeat, slur, time signature etc. successfully to Braille. Park et al. [18] have suggested a method of automatic translation of scanned images of books to digital Braille books. They implemented character identification and recognized images in Books while automatically translating them to text. This method reduced time and cost required for producing books in Braille.

Alufaisan et al. [19] designed an application that identifies Braille numerals in Arabic and converts it to plain text using CNN-based Residual Networks. The system also gave speech assistance to the generated text. Apu et al. [20] proposed user and budget friendly braille device that can translate text and voice to braille for blind students. It works for different languages and converts based on 'text' or 'voice' command given by the user.

Yoo and Baek [21] have proposed a device that can print braille documents for blind. They implemented a raspberry Pi camera to save documents as images stored in device. Characters were extracted from the images and converted to Braille which is then processed to output braille. Their proposed device was portable and could be created using 3D printing. Zhang et al. [22] have used n-gram language model to implement Chinese-braille inter translation system. This system integrates Chinese and Braille word segmentation with concatenation rules. They have also proposed an experimental plan to improve Braille word segmentation and concatenation rules with a word corpus of Chinese-Braille.

III. EXPERIMENTAL SETUP

For development of our machine translation system, we collected English monolingual corpus and translated it into five Indian languages viz Hindi, Marathi, Nepali, Gujarati, and Urdu. Around 1 lac English sentences were collected, and translated by human annotators. The translated sentences were then vetted by another annotator.

After vetting was completed, this corpus was used for training neural machine translation models. For our experiments, we trained two models; first model was a baseline model which had usual preprocessing steps and next was a muti-word expression aware model which along with usual preprocessing steps also had a sub-module to recognize some of the MWEs and also translate them.

Our MWE model could identify compound nouns (e.g., banana shake), light verbs (e.g., book the tickets) and composite name entities (e.g., Narendra Modi).

IV. PROPOSED SYSTEM

For both the machine translation system, we performed some common steps which are the de-facto methods in current state-of-the-art NMT systems. These were:

**Source Text Rewriting:** English has a default structure as Subject-Verb-Object and Indian languages have a default structure as Subject-Object-Verb. Thus, there was a need to transform the structure of English into Indian languages. For this, a transfer engine was developed. In order to process English, we used Stanford CoreNLP library [23]. The transfer engine had 843 handcrafted rules. Snapshot of same is shown in figure 1.

To understand the working of above-mentioned process let's take an example, "Kavita Sharma has booked the ticket for a morning flight to Delhi." Using Stanford parse we will get a parse tree of the sentence (shown in figure 2). This is sent to the transfer engine for conversion to target Indian language (shown in figure 3)

**Sub Wording:** Sub wording is a process of dividing words into smaller units known as sub-words or sub-word units. The objective of sub wording is to capture the meaning of a word by breaking it down into smaller, more basic units that can be more easily understood by machine learning models. These helps substantially in handling out of vocabulary (OOV) words, which are words that are not present in the training data.

The above two process were applied on the entire one lac English sentence corpus. Once this process was completed, we simply trained the baseline NMT model using PyTorch library.

We also trained one more machine translation system which augmented the transfer tree by first recognizing the multiword expressions in the corpus and then translating or transliterating them in the target Indian language. Thus, for our example, let's assume that we wanted to translate the English text into Hindi, then at first, the system would recognize the MWEs in the input sentence. In our example they would be "Kavita Sharma", "booked the ticket" and "morning flights".

TABLE I. DEMONSTRATION OF LAYERED OUTPUT FOR THE EXAMPLE SENTENCE

| | |
|---|---|
| **Input Sentence** | Kavita Sharma has booked the ticket for a morning flight to Delhi. |
| **English Parse in LISP Notation** | [S [NP [NNP Kavita] [NNP Sharma]] [VP [VBZ has] [VP [VBN booked] [NP [NP [DT the] [NN ticket]] [PP [IN for] [NP [NP [DT the] [NN morning] [NN flight]] [PP [TO to] [NP [NNP Delhi]]]]]]]] [. .]] |
| **LISP Notation of Syntax Transferred Tree** | [S [NP [NNP] [NNP] [IN]] [VP [PP [NP [NNP] [TO]]] [PP [NP [NP [NN] [NN]]] [IN]] [VP [VBN] [VBZ]]]] |
| **NE Translated Parse Tree** | [S [NP [NNP कविता] [NNP शर्मा] [IN has]] [VP [PP [NP [NNP Delhi] [TO to]]] [PP [NP [NP [NN सुभा] [NN फ्लाइट]]] [IN for]] [VP [VBN बूक_की] [VBZ हैं]]]] |

Further, the system would translate these MWEs using a separate knowledge base that has been handcrafted by human annotators. In case, if any MWE is not found in the knowledgebase then they get translated. Thus, in our example, the target MWEs would be "कविता शर्मा", "बूक की हैं" and "सुबह की फ्लाइट". These are replaced by the English MWEs in the parse tree. Thus, the resultant parse tree has a mix of English text and Hindi translated text. This is shown in figure 4.

Next the augmented (MWE translated) sentence is sent for sub-wording and then finally is provided to PyTorch library for NMT model training. Figure 5 shows the schematic diagram of the entire working of the system. The working of the entire system is also enumerated in table 2.

Finally, the text translated for the NMT systems was sent the Bharti Braille translation engine which would then translate it into the corresponding Bharti Braille encoding.

## V. EVALUATION

Both the MT systems were tested on one thousand sentences (for each language pair). We used BLEU metric to compare the results of the two systems and found the in all the cases MWE based NMT systems performed better then the baseline NMT systems. The result of this study is shown in table 3. These are system level scores where the individual translation scores were added and were divided by one thousand. The same is shown in equation 1. Here, BLEU-Score$_i$ is the individual BLEU score for each sentence which are then added for all n sentences and are then divided by total number (n) of sentences.

$$System - level\ BLEU\ Score = \frac{\sum_{i=1}^{n} BLEU-Score_i}{n} \quad (1)$$

TABLE II. EVALUATION RESULTS OF NMT MODELS

| Language Pair | Baseline Models | MWE Induced NMT Models | Improvement |
|---|---|---|---|
| **English-Hindi** | 0.5261 | 0.7591 | 23.30% |
| **English-Marathi** | 0.5193 | 0.7489 | 22.96% |
| **English-Nepali** | 0.4937 | 0.7145 | 22.08% |
| **English-Gujarati** | 0.4871 | 0.7433 | 25.62% |
| **English-Urdu** | 0.4693 | 0.6945 | 22.52% |

## VI. CONCLUSION

In this paper, we have shown the development of a MWE augmented NMT system. The proper treatment of multiword expressions in the translation process improves the performance of the system. This is evident through the comparison of results of baseline NMT system with MWE induced NMT system where in all the language pairs there has been some improvement in the results. The performance improvement ranged from 22% to 25% where English-Nepali showed the least improvement with a gain of 22.08% and English-Gujarati showed the most improvement with a gain of 25.62%.

As an extension to this work, we would also like to observe the change in performance of the NMT models where more multi-word expressions are handled in the same manner. This would further strengthen our hypothesis that by adding linguistic knowledge, the performance of the vanilla NMT models can be improved.

ACKNOWLEDGMENT

This work is supported by the funding received from SERB, GoI through grant number CRG/2020/004246 for project entitled, "Development of English to Bharti Braille Machine Assisted Translation System".

| LINK | SOURCE_TREE | TARGET_TREE |
|---|---|---|
| 1.NP_r=2.NP_r.~1.NP_f=2.NP_f.~1.S=2.S.~1.NP_0=2.NP... | nx-ReCl-nx0e-Vpadjn-Vnx1 | nx-ReCl-nx0e-Vpadjn-Vnx1 |
| 1.S_r=2.S_r.~1.S=2.S.~1.NP_0=2.NP_0.~1.VPadjn=2.VP... | nx0e-VPadjn-Vinf-s | snx0e-Vinf |
| 1.PP_r=2.PP_r.~1.Adv=2.Adv.~1.PP_f=2.PP_f.~ | ARBpx | ARBpx |
| 1.NP_r=2.NP_r.~1.NP_f=2.NP_f.~1.S=2.S.~1.NP_0=2.NP... | nx-INF-nx0e-Vpadjn-Vnx1 | nx-INF-nx0e-Vpadjn-Vnx1 |
| 1.S_r=2.S_r.~1.S=2.S.~1.NP_0=2.NP_0.~1.VPadjn=2.VPadjn.~1.ep... | nx0e-VPadjn-Vpx1 | nx0e-px1-V |
| 1.S_r=2.S_r.~1.S=2.S.~1.NP_0=2.NP_0.~1.eps=2.eps.~1... | nx0e-VPadjn-Vinf-nx1nx2-s | snx0e-nx1nx2-Vinf |
| 1.S_r=2.S_r.~1.S=2.S.~1.NP_0=2.NP_0.~1.VPadjn=2.NP_0.~1... | nx0e-VPadjn-Vinf-nx1-s | snx0e-nx1-Vinf |
| 1.S_r=2.S_r.~1.NPadjn=2.NPadjn.~1.NP_0=2.NP.~1.NP... | nx0-VPadjn-Vnx1 | nx0ne-nx1ko-ADJ-vz1 |
| 1.NP_r=2.NP_r.~1.NP_f=2.NP_f.~1.S=2.S.~1.NP_0=2.NP... | nx-nx0-VPadjnnx1 | nx-nx0ne-nx1ko-ADJ-vz1 |
| 1.S_r=2.S_r.~1.NP_0=2.NP_0.~1.NP_1=2.NP.~1.VPadj... | nx0e-VPadjn-Vnx1 | nx0e-nx1ko-ADJ-vz1 |
| 1.NP_r=2.NP_r.~1.NP_f=2.NP_f.~1.S=2.S.~1.NP_0=2.NP... | nx-nx0-VPadjn-V | nx-nx0ne-ADJ-vz1 |
| 1.S_r=2.S_r.~1.NP_0=2.NP_0.~1.VPadjn=2.VPadjn.~1.V... | nx0e-VPadjn-V | nx0e-ADJ-vz1 |
| 1.NP_r=2.NP_r.~1.NP_f=2.NP_f.~1.S=2.S.~1.NP_0=2.NP... | nxsnx0e-VPadjn-V | nxsnx0e-ADJ-vz1 |
| 1.NP_r=2.NP_r.~1.NP_f=2.NP_f.~1.S=2.S.~1.NP_0=2.N... | nxsnx0e-VPadjn-Vnx1 | nxsnx0e-ADJ-vz1 |
| 1.S_r=2.S_r.~1.NPadjn=2.NPadjn.~1.NP_0=2.NP.~1.VP... | nx0-VPadjn-V | nx0ne-ADJ-vz1 |
| 1.S_r=2.S_r.~1.VPadjn=2.VPadjn.~1.NP_0=2.NP_0.~1.V... | nx0-VPadjn-Vnx1nx2 | nx0ne-nx1ko-nx2-ADJ-vz1 |
| 1.VPadjn_r=2.VPadjn_r.~1.PP=2.PP.~1.VPadjn_f=2.VPadj... | PxVPadjn | nxPvxadjn |
| 1.S_r=2.S_r.~1.NPadjn=2.NPadjn.~1.NP_0=2.NP_0.~1.V... | nx0-VPadjn-VP-AdjP | nx0-AdjP-V |
| 1.S=2.S_r.~1.S_1=2.S_1.~1.Cond=2.Cond.~1.S_2=2.S_2.~ | S1CondS2 | S1CondS2 |
| 1.S_r=2.S_r.~1.NP_0=2.NP_0.~1.VPadjn=2.VPadjn.~1.ep... | nx0e-VPadjn-Vnx1nx2 | nx0e-nx1ko-nx2-V |
| 1.S_r=2.S_r.~1.NPadjn=2.NPadjn.~1.NP_0=2.NP.~1.NP... | nx0-VPadjn-Vnx1 | nx0ne-nx1ko-ADJ-vz1 |
| 1.NP_r=2.NP_r.~1.NP_f=2.NP_f.~1.S=2.S.~1.NP_0=2.NP... | nx-nx0-VPadjnnx1 | nx-nx0ne-nx1ko-ADJ-vz1 |
| 1.S_r=2.S_r.~1.NP_0=2.NP_0.~1.NP_1=2.NP.~1.VPadj... | nx0e-VPadjn-Vnx1 | nx0e-nx1ko-ADJ-vz1 |
| 1.NP_r=2.NP_r.~1.NP_f=2.NP_f.~1.S=2.S.~1.NP_0=2.NP... | nx-nx0-VPadjn-V | nx-nx0ne-ADJ-vz1 |
| 1.S_r=2.S_r.~1.NP_0=2.NP_0.~1.VPadjn=2.VPadjn.~1.V... | nx0e-VPadjn-V | nx0e-ADJ-vz1 |
| 1.NP_r=2.NP_r.~1.NP_f=2.NP_f.~1.S=2.S.~1.NP_0=2.NP... | nxsnx0e-VPadjn-V | nxsnx0e-ADJ-vz1 |
| 1.NP_r=2.NP_r.~1.NP_f=2.NP_f.~1.S=2.S.~1.NP_0=2.N... | nxsnx0e-VPadjn-Vnx1 | nxsnx0e-nx1-ADJ-vz1 |

**Fig. 1.** Snapshot of transfer grammar rules.

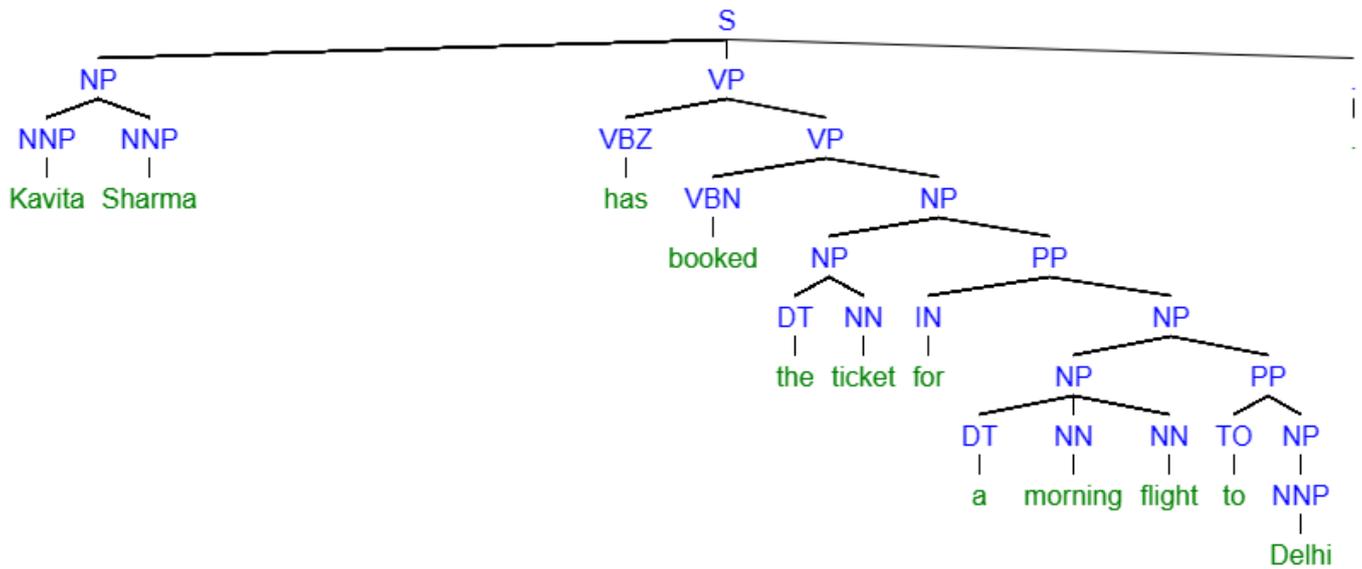

FIGURE 2    EXAMPLE PARSE OF ENGLISH SENTENCE.

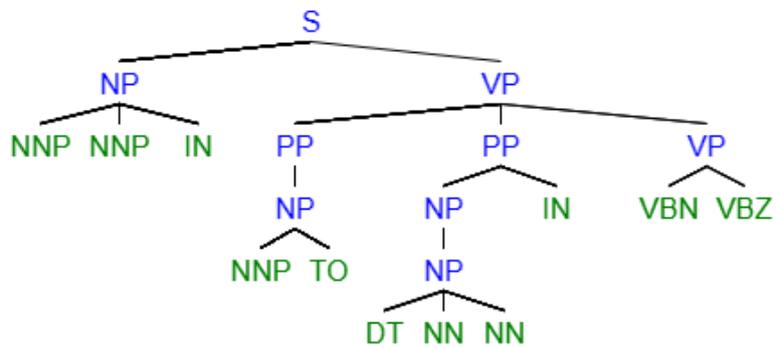

FIGURE 3    EXAMPLE ENGLISH PARSE TREE WITHOUT LEAF NODES

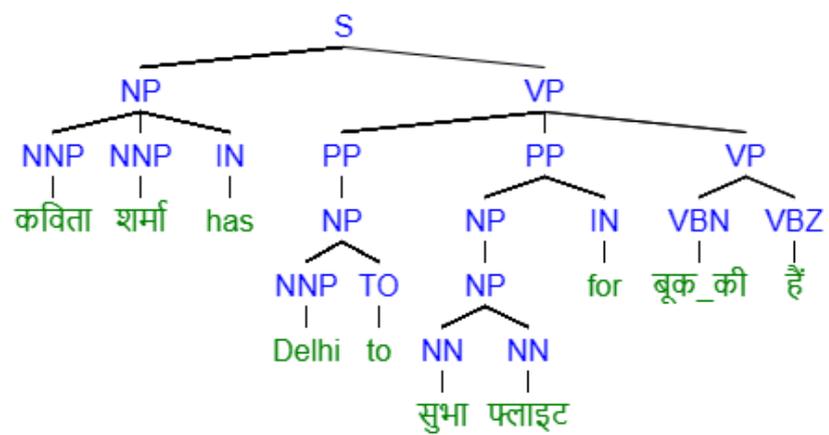

FIGURE 4    EXAMPLE PARSE TREE WITH HINDI MWEs

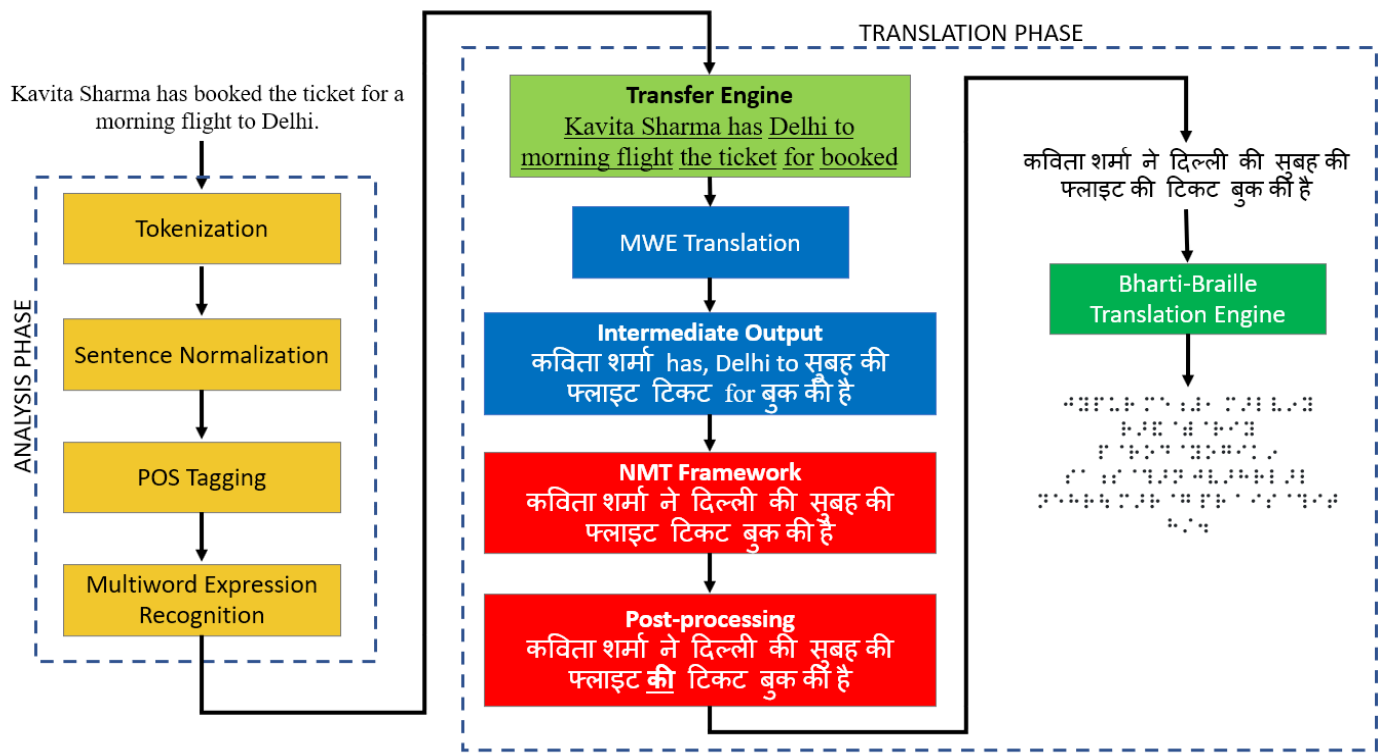

FIGURE 5 WORKFLOW OF MWE RECOGNITION BASED NEURAL MACHINE TRANSLATION SYSTEM.